\documentclass[runningheads]{llncs}
\usepackage{graphicx}
\graphicspath{ {./image/} }

\usepackage{orcidlink}
\usepackage{nth}
\renewcommand{\orcidID}[1]{\orcidlink{#1}}

\usepackage{hyperref}
\hypersetup{hidelinks,
breaklinks=true,
colorlinks=true,%
urlcolor=blue,
bookmarksopen=false,
pdftitle={ICDAR 2021 Competition on Historical Map Segmentation},
pdfauthor={J. Chazalon et al.}}

\usepackage{amsmath}
\usepackage{booktabs}

\usepackage{caption}
\usepackage{subcaption}
\usepackage[inline]{enumitem}

\usepackage{cleveref}

\begin{document}
\title{ICDAR 2021 Competition on\\Historical Map Segmentation%
\thanks{This work was partially funded by the French National Research Agency (ANR):
Project SoDuCo, grant ANR-18-CE38-0013.
We thank the City of Paris for granting us with the permission to use and reproduce the atlases used in this work.}}
\titlerunning{ICDAR'21 MapSeg}
\author{%
Joseph Chazalon\inst{1}\orcidID{0000-0002-3757-074X}\and
Edwin Carlinet\inst{1}\orcidID{0000-0001-5737-5266}\and
Yizi Chen\inst{1,2}\orcidID{0000-0003-1637-0092}\and
Julien Perret\inst{2,3}\orcidID{0000-0002-0685-0730}\and
Bertrand Duménieu\inst{3}\orcidID{0000-0002-2517-2058}\and
Clément Mallet\inst{2}\orcidID{0000-0002-2675-165X}\and
Thierry Géraud\inst{1}\orcidID{0000-0002-0380-7948}\and\\
Vincent Nguyen\inst{4,5}\orcidID{0000-0003-2271-6918}\and
Nam Nguyen\inst{4}\and\\ %
Josef Baloun\inst{6,7}\orcidID{0000-0003-1923-5355}\and
Ladislav Lenc\inst{6,7}\orcidID{0000-0002-1066-7269}\and
Pavel Král\inst{6,7}\orcidID{0000-0002-3096-675X}
}
\authorrunning{J. Chazalon et al.}
\institute{%
EPITA Research and Development Lab. (LRDE), EPITA, France \and
Univ. Gustave Eiffel, IGN-ENSG, LaSTIG, France \and
LaDéHiS, CRH, EHESS, France\and
L3i, University of La Rochelle, France\and
Liris, INSA-Lyon, France\and
Department of Computer Science and Engineering, University of West Bohemia, Univerzitní, Pilsen, Czech Republic\and
NTIS - New Technologies for the Information Society, University of West Bohemia, Univerzitní, Pilsen, Czech Republic%
}
\maketitle              %
\begin{abstract}
This paper presents the final results of the ICDAR 2021 Competition on Historical Map Segmentation (MapSeg),
encouraging research on a series of historical atlases of Paris, France, drawn at 1/5000 scale between 1894 and 1937.
The competition featured three tasks, awarded separately.
Task~1 consists in detecting building blocks and was won by the L3IRIS team using a DenseNet-121 network trained in a weakly supervised fashion. This task is evaluated on 3 large images containing hundreds of shapes to detect.
Task~2 consists in segmenting map content from the larger map sheet, and was won by the UWB team using a U-Net-like FCN combined with a binarization method to increase detection edge accuracy.
Task~3 consists in locating intersection points of geo-referencing lines, and was also won by the UWB team who used a dedicated pipeline combining binarization, line detection with Hough transform, candidate filtering, and template matching for intersection refinement.
Tasks~2 and~3 are evaluated on 95 map sheets with complex content.
Dataset, evaluation tools and results are available under permissive licensing at \url{https://icdar21-mapseg.github.io/}.

\end{abstract}
\section{Introduction}
\begin{figure*}[tb]
    \centering
    \includegraphics[width=.95\linewidth]{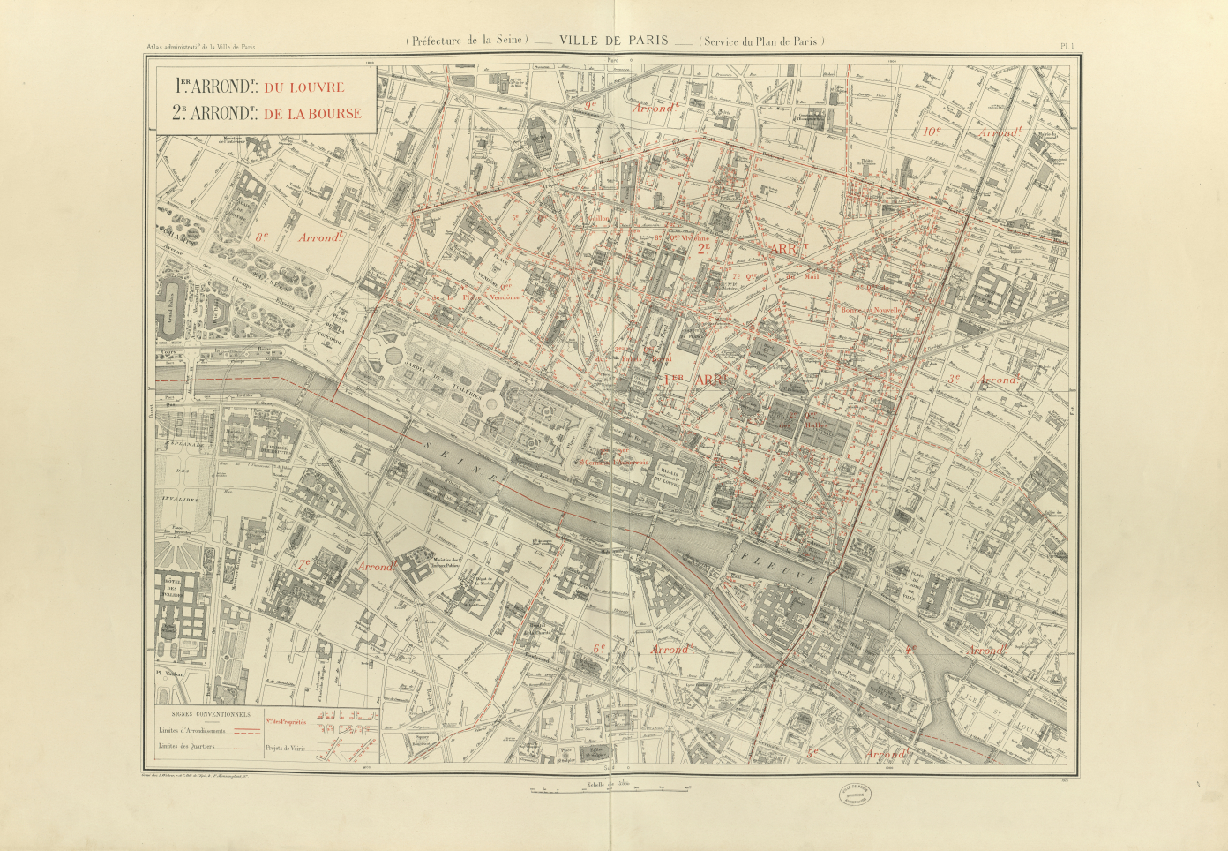}
    \caption{Sample map sheet. Original size: 11136$\times$7711 pixels.} %
    \label{fig:map_large_in}
\end{figure*}
\begin{figure}[tb]
    \centering
    \begin{subfigure}[b]{0.48\textwidth}
        \centering
        \includegraphics[width=\linewidth,trim={206px 217px 40px 195px},clip]{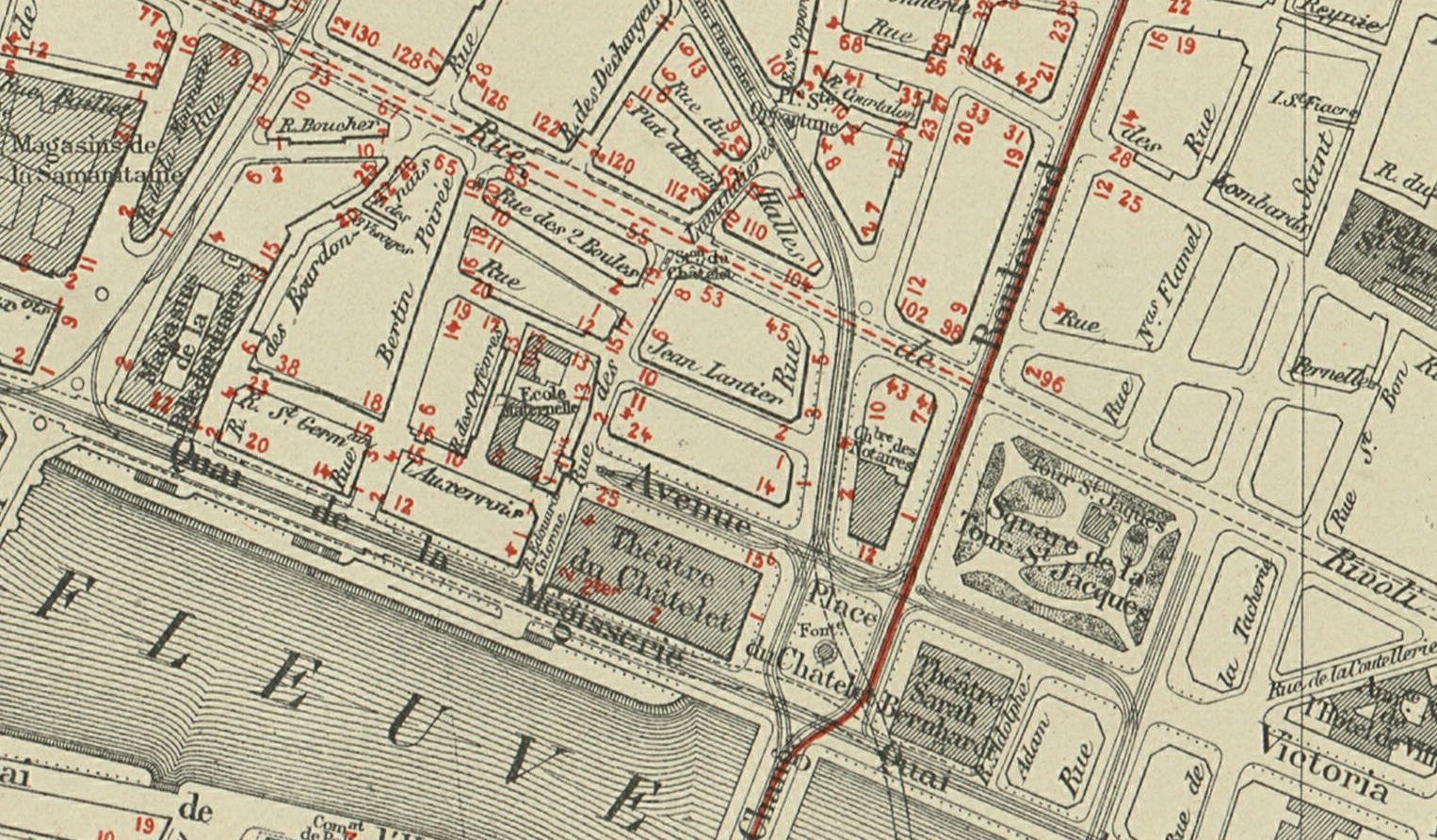}
    \end{subfigure}
    \hfill
    \begin{subfigure}[b]{0.48\textwidth}
        \centering
        \includegraphics[width=\linewidth,trim={0 20px 22px 60px},clip]{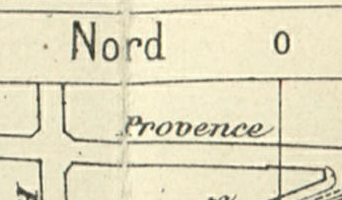}
    \end{subfigure}
    \caption{Some map-related challenges (\emph{left}): visual polysemy, planimetric overlap, text overlap\dots{} and some document-related challenges (\emph{right}): damaged paper, non-straight lines, image compression, handwritten text\dots}
    \label{fig:challenges}
\end{figure}
\textbf{Motivation~}
This competition consists in solving several challenges which arise during the processing of images of historical maps.
In the Western world, the rapid development of geodesy and cartography from the $18^{th}$ century resulted in massive production of topographic maps at various scales.
City maps are of utter interest. They contain rich, detailed, and often geometrically accurate representations of numerous geographical entities. 
Recovering spatial and semantic information represented in old maps requires a so-called \textit{vectorization} process. 
Vectorizing maps consists in transforming rasterized graphical representations of geographic entities (often maps) into instanced geographic data (or vector data), that can be subsequently manipulated (using Geographic Information Systems). 
This is a key challenge today to better preserve, analyze and disseminate content for numerous spatial and spatio-temporal analysis purposes.

\textbf{Tasks~}
From a document analysis and recognition (DAR) perspective, full map vectorization covers a wide range of challenges, from content separation to text recognition.
\emph{MapSeg} focuses on 3 critical steps of this process.
\begin{enumerate*}[label=(\roman*)]
    \item Hierarchical information extraction, and more specifically the \emph{detection of building blocks} which form a core layer of map content.
    This is addressed in \emph{Task~1}.
    \item \emph{Map content segmentation}, which needs to be performed very early in the processing pipeline to separate meta-data like titles, legends and other elements from core map content.
    This is addressed in \emph{Task~2}.
    \item Geo-referencing, i.e. mapping historical coordinate system into a recent reference coordinate system.
    Automating this process requires the detection of well-referenced points, and we are particularly interested in detecting \emph{intersection points between graticule lines} for this purpose.
    This is addressed in \emph{Task~3}.
\end{enumerate*}

\textbf{Dataset~}
\emph{MapSeg} dataset is extracted from a series of 9 atlases of the City of Paris\footnote{
\textit{Atlas municipal des vingt arrondissements de Paris}.
1894, 1895, 1898, 1905, 1909, 1912, 1925, 1929, and 1937.
\textit{Bibliothèque de l’Hôtel de Ville}.
City of Paris. France.
Online resources for the 1925 atlas: \url{https://bibliotheques-specialisees.paris.fr/ark:/73873/pf0000935524}.
} produced between 1894 and 1937 by the Map Service (\emph{``Service du plan''}) of the City of Paris, France, for the purpose of urban management and planning.
For each year, a set of approximately 20 sheets forms a tiled view of the city, drawn at 1/5000 scale using trigonometric triangulation.
\Cref{fig:map_large_in} is an example of such sheet.
Such maps are highly detailed and very accurate even by modern standards.
This material provides a very valuable resource for historians and a rich body of scientific challenges for the document analysis and recognition (DAR) community: map-related challenges (\cref{fig:challenges}, \emph{left}) and document-related ones (\cref{fig:challenges}, \emph{right}).
The actual dataset is built from a selection of 132 map sheets extracted from these atlases. 
Annotation was performed manually by trained annotators and required 400 hours of manual work to annotate the 5 sheets of the dataset for Task~1 and 100 hours to annotate the 127 sheets for Tasks~2 and~3.
A special care was observed to minimize the spatial and, to some extent, temporal overlap between train, validation and test sets:
except for Task~1 which contains districts 1 and 2 both in train and test sets (from different atlases to assess the potential for the exploitation of information redundancy over time), each district is either in the training, the validation or the test set.
This is particularly important for Tasks~2 and~3 as the general organization of the map sheets are very similar for sheets representing the same area.

\textbf{Competition Protocol~}
The MapSeg challenge ran from November 2020 to April 2021.
Participants were provided with training and validation sets for each 3 tasks by November 2020.
The test phase for Task~1 started with the release of test data on April 5$^{th}$, 2021 and ended in April 9$^{th}$.
The test phase for Tasks~2 and~3 started with the release of test data on April 12$^{th}$, 2021 and ended in April 16$^{th}$.
Participants were requested to submit the results produced by their methods (at most 2) over the test images,
computed by themselves.

\textbf{Open Data, Tools and Results~}
To improve the reliability of the evaluation and competition transparency, evaluation tools were released (executable and open source code) early in the competition, so participants were able to evaluate their methods on the validation set using exactly the same tools as the organizers on the test set.
More generally, all competition material is made available under very permissive licenses at \url{https://icdar21-mapseg.github.io/}:
dataset with ground truth, evaluation tools and participants results can be downloaded, inspected, copied and modified.

\vspace{0.5em}
\textbf{Participants~}
The following teams took part in the competition.
\vspace{-0.5em}
\begin{description}
    \item[CMM Team:] 
    \textit{Team:} Mateus Sangalli, Beatriz Marcotegui, Jos\'e Marcio Martins Da Cruz, Santiago Velasco-Forero, Samy Blusseau;
    \textit{Institutions:} Center for Mathematical Morphology, Mines ParisTech, PSL Research University, France;
    \textit{Extra material:} 
    morphological library Smil: \url{http://smil.cmm.minesparis.psl.eu};
    code: \url{https://github.com/MinesParis-MorphoMath}.
    \item[IRISA Team:] 
    \textit{Competitor:} Aurélie Lemaitre;
    \textit{Institutions:} IRISA/Université Rennes 2, Rennes, France;
    \textit{Extra material:}  \url{http://www.irisa.fr/intuidoc/}.
    \item[L3IRIS Team:] 
    \textit{Team:} Vincent Nguyen and Nam Nguyen;
    \textit{Institutions:} L3i, University of La Rochelle, France; Liris, INSA-Lyon, France;
    \textit{Extra material:} \url{https://gitlab.univ-lr.fr/nnguye02/weakbiseg}.
    \item[UWB Team:] 
    \textit{Team:} Josef Baloun, Ladislav Lenc and Pavel Král;
    \textit{Institutions:}
    Department of Computer Science and Engineering, University of West Bohemia, Univerzitní, Pilsen, Czech Republic;
    NTIS - New Technologies for the Information Society, University of West Bohemia, Univerzitní, Pilsen, Czech Republic;
    \textit{Extra material:} \url{https://gitlab.kiv.zcu.cz/balounj/21_icdar_mapseg_competition}.
    \item[WWU Team:] 
    \textit{Team:} Sufian Zaabalawi, Benjamin Risse;
    \textit{Institution:} Mün\-ster University, Germany;
    \textit{Extra material:} Project on automatic vectorization of historical cadastral maps: \url{https://dhistory.hypotheses.org/346}.
\end{description}

\section{Task~1: Detect Building Blocks}
\begin{figure}[tb]
    \centering
    \begin{subfigure}[b]{0.45\textwidth}
        \centering
        \includegraphics[width=\linewidth,clip]{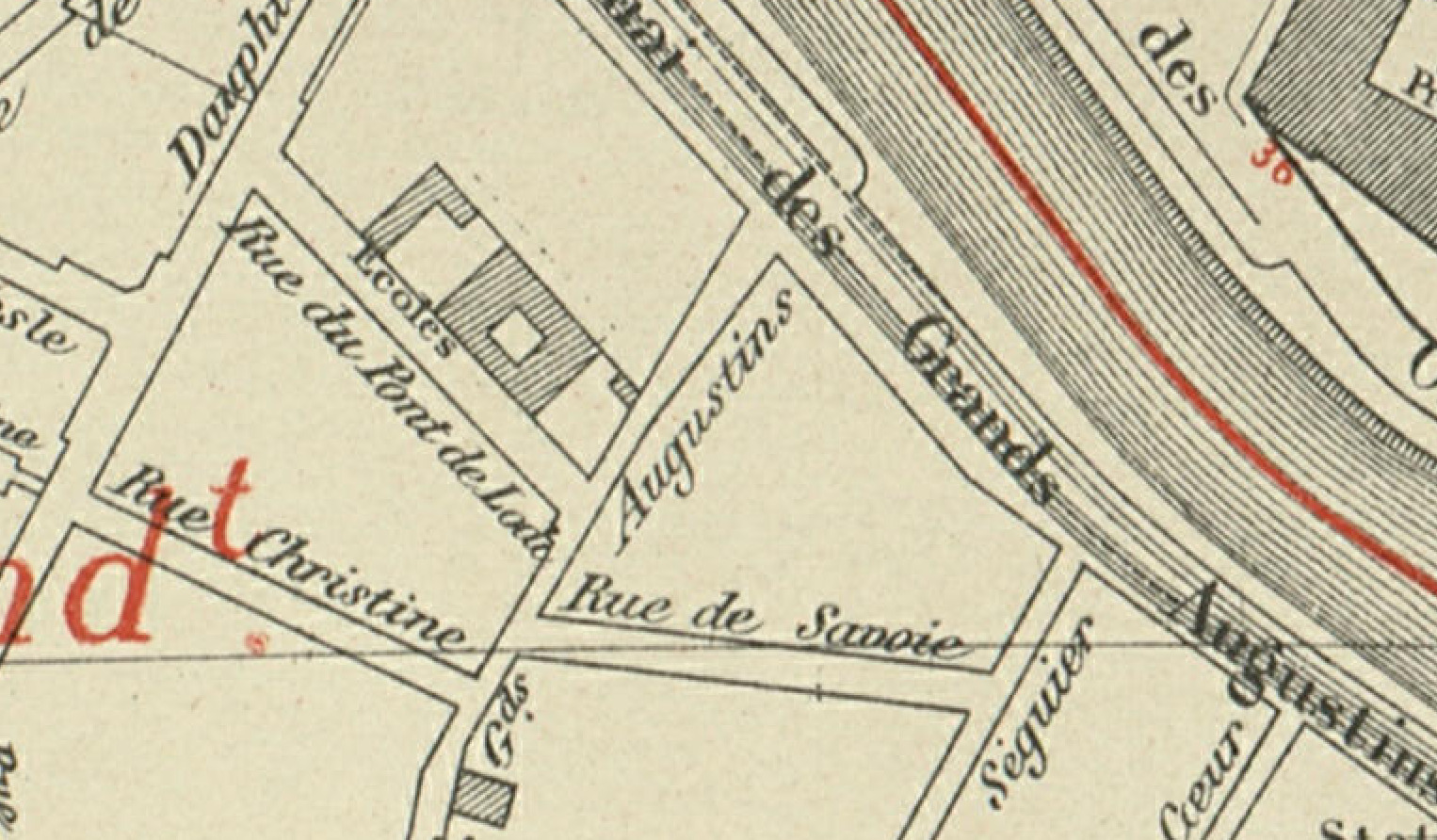}
    \end{subfigure}
    \hfill
    \begin{subfigure}[b]{0.45\textwidth}
        \centering
        \includegraphics[width=\linewidth,clip]{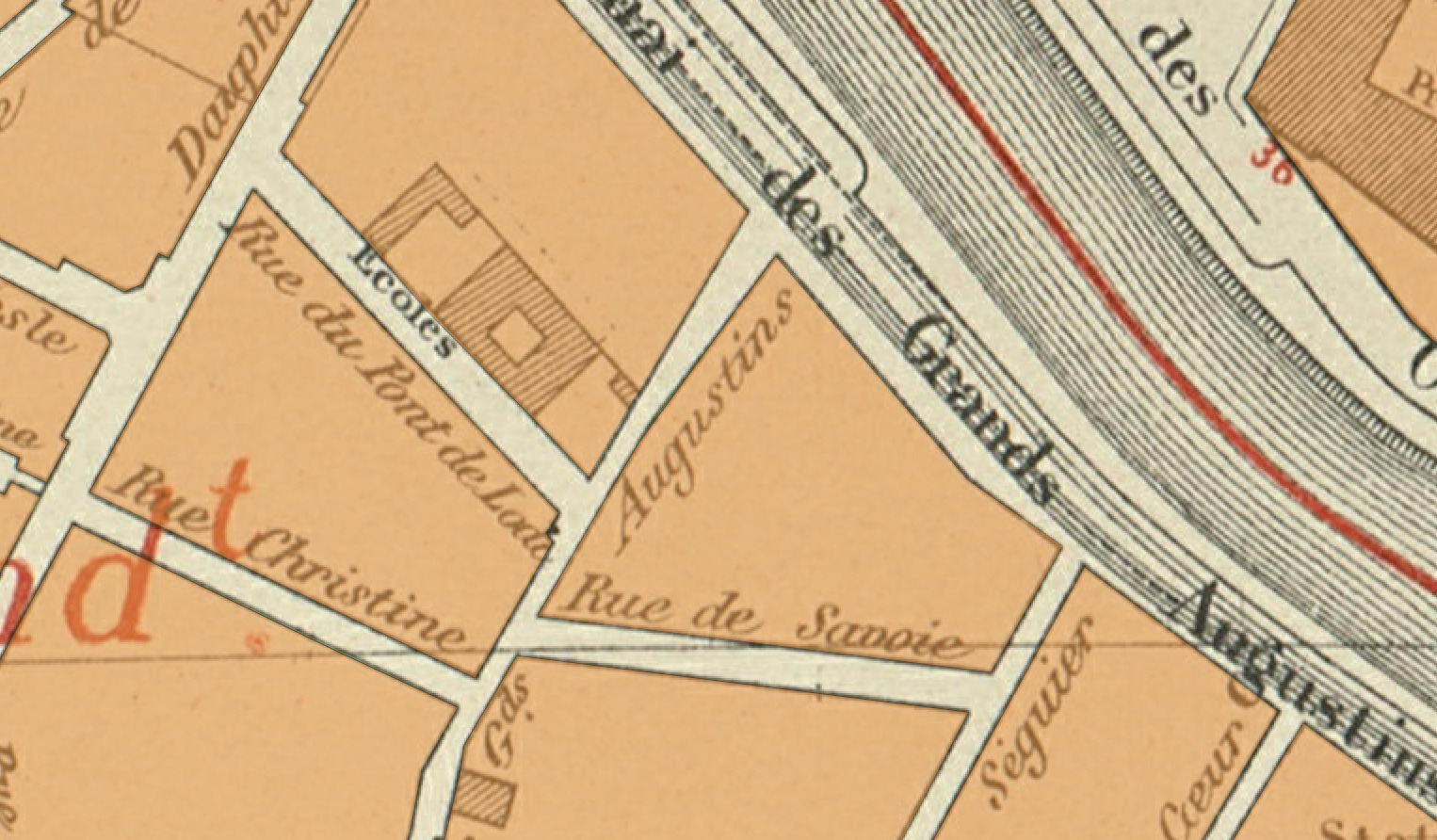}
    \end{subfigure}
    \caption{Input (\emph{left}) and output (\emph{right}, orange overlay) image excerpts for task~1.}
    \label{fig:task1-inout}
\end{figure}

This task consists in detecting a set of closed shapes (building blocks) in the map image.
Building blocks are coarser map objects which can regroup several elements.
Detecting these objects is a critical step in the digitization of historical maps because it provides essential components of a city.
Each building block is symbolized by a closed shape which can enclose other objects and lines.
Building blocks are surrounded by streets, rivers fortification wall or others, and are never directly connected.
Building blocks can sometimes be reduced to a single spacial building, symbolized by diagonal hatched areas.
Given the image of a complete map sheet and a mask of the map area, participants had to detect each building block as illustrated in \cref{fig:task1-inout}.

\subsection{Dataset}
Inputs form a set of JPEG RGB images which are quite large ($\approx8000\times8000$ pixels).
They are complete map sheet images as illustrated in \cref{fig:map_large_in}, cropped to the relevant area (using the ground truth from Task~2). Remaining non-relevant pixels were replaced by black pixels.
Expected output for this task is a binary mask of the building blocks: pixels falling inside a building block must be marked with value 255, and other pixels must be marked with value 0.
The dataset is separated as follows:
the train set contains 1 image (sheet 1 of 1925 atlas --- 903 building blocks),
the validation set contains 1 image (sheet 14 of 1925 atlas --- 659 building blocks),
and the test set contains 3 images (sheet 1 of 1898 atlas, sheet 3 of 1898 and 1925 atlases --- 827, 787 and 828 building blocks).

\subsection{Evaluation Protocol}
Map vectorization requires an accurate detection of shape boundaries.
In order to assess both the detection and segmentation quality of the target shapes, and therefore to avoid relying only on pixel-wise accuracy, we use the COCO Panoptic Quality (PQ) score~\cite{kirillov2019panoptic} which is based on an instance segmentation metric.
This indicator is computed as follows: 
$$
{\text{PQ}} = \underbrace{\frac{\sum_{(p, g) \in TP} \text{IoU}(p, g)}{\vphantom{\frac{1}{2}}|TP|}}_{\text{segmentation quality (SQ)}} \times \underbrace{\frac{|TP|}{|TP| + \frac{1}{2} |FP| + \frac{1}{2} |FN|}}_{\text{recognition quality (RQ)}}
$$
where $TP$ is the set of matching pairs $(p, g) \in (P \times G)$ between predictions ($P$) and reference ($G$), $FP$ is the set of unmatched predicted shapes, and $FN$ is the set of unmatched reference shapes. 
Shapes are considered as matching when:
$$
\text{IoU}(p,g) = \frac{p \cap g}{p \cup g} > 0.5.
$$
\textbf{COCO SQ} (segmentation quality) is the mean IoU between matching shapes: matching shapes in reference and prediction have an IoU $>$ 0.5.
\textbf{COCO RQ} (detection/recognition quality) is detection F-score for shape: a predicted shape is a true positive if it as an IoU $>$ 0.5 with a reference shape. 
The resulting \textbf{COCO PQ} indicator ranges from 0 (worst) to 1 (perfect).
An alternate formulation of this metric is to consider it as the integral of the detection F-score over all possible IoU thresholds between 0 and 1~\cite{chazalon21.icdar}.
We also report the ``F-Score vs IoU threshold'' curves for each method to better assess their behavior.

\subsection{Method descriptions}

\subsubsection{L3IRIS --- \emph{Winning Method}}
\label{sec:task1-l3iris}
From previous experiments, the L3IRIS team reports that both semantic segmentation and instance segmentation approaches led to moderate results, mainly because of the amount of available data.
Instead of trying to detect building blocks directly, they reversed the problem and tried to detect non-building-blocks as a binary semantic segmentation problem which facilitated the training process.
Additionally, they considered the problem in a semi/weakly supervised setting where training data includes 2 images (val+train): the label is available for one image and is missing for the other image.
Their binary segmentation model relies on a DenseNet-121~\cite{huang2017densely} backbone and is trained using the weakly supervised learning method from~\cite{nguyen2020learning}.
Input is processed as 480$\times$480 tiles.
During training, tiles are generated by random crops and augmented with standard perturbations: noise, rotation, gray, scale, etc.
Some post-processing techniques like morphological operators are used to refine the final mask (open the lightly connected blocks and fill holes). The parameter for the post-processing step is selected based on the results on the validation image. Training output is a single models; not ensemble technique is used.
The weakly supervised training enables the training of a segmentation network in a setting where only  a subset of the training objects are labelled.
The overcome the problem of missing labels, the L3IRIS team uses two energy maps $w+$ et $w-$ which represent the certainty that a pixel being positive (object’s pixel) or negative (background pixel). These two energy maps are evolved during the training at each epoch based on prediction scores in multiple past epochs. This technique implicitly minimizes the entropy of the predictions on unlabeled data which can help to boost the performance of the model~\cite{grandvalet2004semi}.
For Task~1, they considered labeled images (from training set) and unlabeled ones (from validation set) for which the label was retained for validation. The two energy maps are applied only for images from the validation set.

\subsubsection{CMM}
\textbf{Method 1}
This method combines two parallel sub-methods, named \textbf{1A} and \textbf{1B}. In both sub-methods, the RGB images are first processed by a U-Net \cite{ronneberger2015u} trained to find edges of the blocks. Then, a morphological processing is applied to the image of edges in order to segment the building blocks. Finally, the results of both sub-methods are combined to produce a unique output.
In method \textbf{1A}, the U-Net is trained with data augmentation that uses random rotations, reflections, occlusion by vertical and horizontal lines and color changes.
The loss function penalizes possible false positive pixels (chosen based on the distance from edge pixels in the color space) with a higher cost.
To segment the blocks, first a top-hat operator is applied to remove large vertical and horizontal lines and an anisotropic closing~\cite{blusseau18tropical} is used to remove holes in the edges.
Edges are binarized by a hysteresis threshold and building blocks are obtained by a fill-holes operator followed by some post-processing. The result is a clean segmentation, but with some relatively large empty regions.
In method \textbf{1B}, the U-Net is trained using the same augmentation process, but the loss function weights errors to compensate class imbalance.
Method 2 (described hereafter) is applied to the inverse of the edges obtained by this second U-Net.
To combine both methods, a closing by a large structuring element is used to find the dense regions in the output of \textbf{1A}, the complement of which are the empty regions of \textbf{1A}.
The final output is equal to \textbf{1A} in the dense region and to \textbf{1B} in the empty regions.

\textbf{Method 2}
This method is fully morphological. It works on a gray-level image, obtained by computing the luminance from the RGB color image. The main steps are the following:
\begin{enumerate*}[label=(\roman*)]
    \item \emph{Area closing} with 1000 pixels as area threshold; this step aims at \emph{removing small dark components} such as letters.
    \item \emph{Inversion of contrast} of the resulting image followed by a \emph{fill-holes} operator; %
    The result of this step is an image where most blocks appear as flat zones, and so do large portions of streets.
    \item The \emph{morphological gradient} of the previous image is computed and then \emph{reconstructed by erosion} starting from markers chosen as the minima with dynamic not smaller than $h = 2$. This produces a new, simplified, gradient image.
    \item The \emph{watershed transform} is applied to the latter image, yielding a labelled image.
    \item A \emph{filtering of labelled components} removes (that is, sets to zero) components smaller than 2000 pixels and larger than 1M pixels; and also removes the components with area smaller than $0.3$ times the area of their bounding box (this removes large star-shaped street components). All kept components are set to 255 to produce a binary image.
    \item A \emph{fill-holes} operator is applied to the latter binary image.
    \item \emph{Removal of river components}: A mask of the river is computed by a morphological algorithm based on a large opening followed by a much larger closing, a threshold and further refinements. Components with an intersection of at least 60\% of their area with the river mask are removed.
\end{enumerate*}

\subsubsection{WWU}
\textbf{Method 1}
This method relies on a binary semantic segmentation U-Net~\cite{ronneberger2015u} trained with Dice loss on a manual selection of 2000 image patches of size 256$\times$256$\times$3 covering representative map areas.
The binary segmentation ground truth is directly used as target.
This U-Net architecture uses 5 blocks and each convolutional layer has a 3$\times$3 kernel filter with no padding. The expending path uses transpose convolutions with 2$\times$ stride. Intermediate activations use RELU and the final activation is a sigmoid.

\textbf{Method 2}
This method relies on a Deep Distance transform.
The binary label image is first transformed into a Euclidean \emph{distance transform map} (DTM),
which highlights the inner region of the building blocks more than the outer edges.
Like for method 1, 2000 training patch pairs of size 256$\times$256 are generated.
The U-Net architecture is similar to the previous method, except that the loss function is the mean square error loss (MSE) and the prediction is a DTM.
This DTM is then thresholded and a watershed transform is used to fill regions of interest and extract building block shapes.

\subsection{Results and discussion}
\begin{figure}[tb]
    \centering
    \begin{subfigure}{0.48\textwidth}
        \centering\begin{tabular}{clc}
            \toprule
            Rank & Team (method) & COCO PQ (\%) $\uparrow$\\
            \midrule
            1 & L3IRIS & 74.1\\
            2 & CMM (1) & 62.6\\
            3 & CMM (2) & 44.0\\
            4 & WWU (1) & 06.4\\
            5 & WWU (2) & 04.2\\
            \bottomrule
        \end{tabular}
    \end{subfigure}
    \hfill
    \begin{subfigure}{0.48\textwidth}
        \centering
        \includegraphics[width=\linewidth,trim={10px 35px 15px 45px},clip]{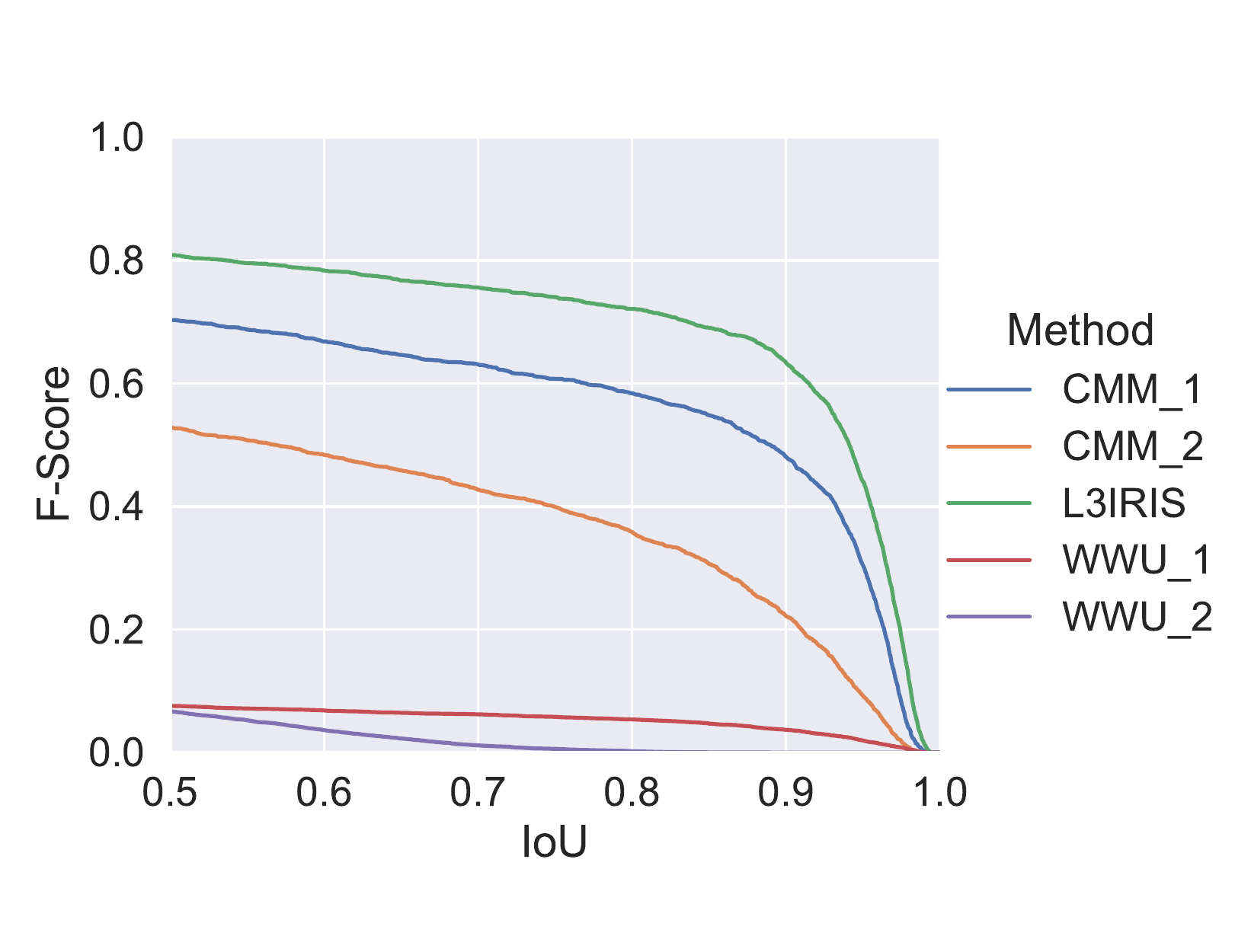}
    \end{subfigure}
    \caption{Final COCO PQ scores for Task~1 (\emph{left}) --- score ranges from 0 (worst) to 100\% (best),
    and plot of the F-Score for each threshold between 0.5 and 1 (\emph{right}). COCO PQ is the area under this curve plus $\frac{1}{2}$ times the intersect at 0.5.}
    \label{fig:task1-results}
\end{figure}
Results and ranking of each submitted method is given in~\cref{fig:task1-results}:
table on the left summarizes the COCO PQ indicators we computed
and the plot on the right illustrates the evolution of the retrieval F-Score for each possible IoU threshold between 0.5 and 1.
While the WWU 1 and 2 approaches get low scores, it must be noted that this is due to local connections between shapes which led to a strong penalty with the COCO PQ metric.
As they implement very natural ways to tackle such problem, they provide a very valuable performance baseline.
CMM method 2 is fully morphological and suffers from an insufficient filtering of noise content,
while CMM method 1 gets much better results by combining edge filtering using some fully convolutional network with morphological filtering.
L3IRIS winning approach improves over this workflow by leveraging weakly-supervised training followed by some morphological post-processing.

\section{Task~2: Segment Map Content Area}
\begin{figure}[tb]
    \centering
    \includegraphics[width=.75\linewidth,clip]{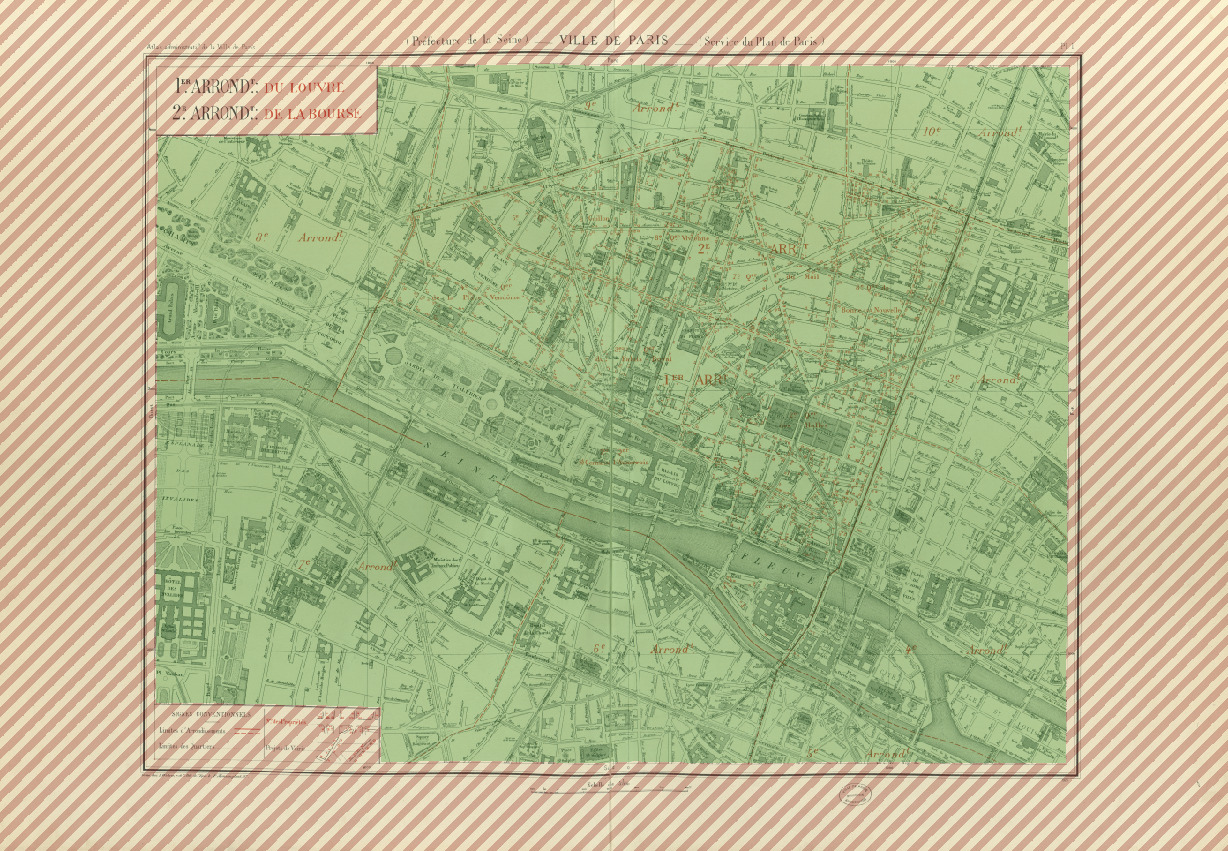}
    \caption{Illustration of expected outputs for Task~2: green area is the map content and red hatched area is the background.}
    \label{fig:task2-output-large}
\end{figure}

This task consists in segmenting the map content from the rest of the sheet.
This is a rather classical document analysis task as it consists in focusing on the relevant area in order to perform a dedicated analysis. In our case, Task~1 would be the following stage in the pipeline.
Given the image of a complete map sheet (illustrated in \cref{fig:map_large_in}), participants had to locate the boundary of the map content, as illustrated in \cref{fig:task2-output-large}.

\subsection{Dataset}
The inputs form a set of JPEG RGB images which are quite large ($\approx 10000\times10000$ pixels).
They are complete map sheet images.
Expected output for this task is a binary mask of the map content area: pixels belonging to map contents are marked with value 255 and other pixels are marked with value 0.
The dataset is separated as follows:
the train set contains 26 images,
the validation set contains 6 images,
and the test set contains 95 images.
As in the atlas series we have multiple occurrences of similar sheets representing the same area,
we made sure that each sheet appeared in only one set.

\subsection{Evaluation Protocol}
We evaluated the quality of the segmentation using the Hausdorff distance between the ground-truth shape and the predicted one.
This measure has the advantage over the IoU, Jaccard index and other area measures that it keeps a good ``contrast'' between results in the case of large objects (because there is no normalization by the area).
More specifically, we use the ``Hausdorff 95'' variant which discards the 5 percentiles of higher values (assumed to be outliers) to produce a more stable measure.
Finally, we averaged indicators for all individual map images to produce a global indicator.
The resulting measure is an error measure which ranges from 0 (best) to a large value depending on image size.

\subsection{Method descriptions}

\subsubsection{UWB --- \emph{Winning Method}}
\label{sec:task2-uwb}
This method is based on the U-Net-like fully convolutional networked used in~\cite{chronseg}.
This network takes a whole downsampled page as an input and predicts the border of the expected area.
Border training samples were generated from the original ground truth (GT) files.
The values are computed using a Gaussian function, where $\sigma = 50$ and the distance from border is used.
Training is performed using the training set and augmentation (mirroring, rotation and random distortion)~\cite{augmentor} to increase the amount of the training samples.
To improve the location of the detected border, a binarized image is generated using a recursive Otsu filter~\cite{nina2011recursive} followed by the removal of small components.
Network prediction and the binarized image are post-processed and combined with the use of connected components and morphological operations, which parameters are calibrated on the validation set.
The result of the process is the predicted mask of a map content area.

\subsubsection{CMM}
This method assumes that the area of interest is characterized by the presence of black lines connected to each other all over the map.
These connections being due either to the external frame, graticule lines, streets, or text superimposed to the map.
Thus, the main idea is detecting black lines and reconstructing them from a predefined rectangle marker in the center of the image.
The method is implemented with the following steps.
\begin{enumerate*}[label=(\roman*)]
\item Eliminate map margin ($M$) detected by the quasi-flat zone algorithm and removed from the whole image ($I_0$):
  $I=I_0 - M$.
\item Then, black lines ($B$), are extracted by a thresholded black-top-hat ($B=Otsu(I-\varphi(I))$)
\item A white rectangle ($R$) is drawn centered in the middle of the image and of dimensions
  $\frac{W}{2} \times \frac{H}{2}$, with $W$ and $H$ the image width and height respectively.
\item Black lines $B$ are reconstructed by dilation from the centered rectangle: $B_s=Build(R,B)$. Several dark frames
  surround the map. Only the inner one is connected to the drawn rectangle $R$ and delimits the area of interest.
\item The frame (black line surrounding the map) may not be complete due to lack of contrast or noise. A watershed is
  applied to the inverse of the distance function in order to close the contour %
  with $markers= R \, \cup \, border$. The region corresponding to marker $R$ becomes the area of interest.
\item Finally, legends are removed from the area of interest. Legends are the regions from the inverse of
  $B_s$ that are rectangular and close to the border of the area of interest.
\end{enumerate*}

\subsubsection{IRISA}
\label{sec:task2-irisa}
This approach is entirely based on a grammatical rule-based system~\cite{lemaitre08} which combines visual clues of line
segments extracted in the document at various levels of a spatial pyramid. The line segment extractor, based on
a multi-scale Kalman filtering~\cite{leplumey95}, is fast and robust to noise, and can deal with slope and
curvature.
Border detection is performed in two steps:
\begin{enumerate*}[label=(\roman*)]
\item Line segments are first extracted at low resolution level (scale 1:16) to provides visual
  clues on the presence of the double rulings outside the map region.  Then, the coarse enclosing rectangle is detected
  using a set of grammar rules.
\item Line segments are extracted at medium resolution level (scale 1:2) to detect parts of
  contours of the maps, and of title regions. Another set of grammar rules are then used to describe and detect the
  rectangular contour with smaller rectangle (title and legend) in the corners.
\end{enumerate*}
The performance is this approach is limited by the grammar rules which do not consider, though it would be possible, the fact that the map content may lay outside the rectangular region, nor it considers that some legend components may not be located at the corners of the map.

\subsubsection{L3IRIS}
\label{sec:task2-l3iris}
This approach leverages the cutting edge few-shot learning technique HSNet~\cite{min2021hypercorrelation} for image segmentation, in order to cope with the limited amount of training data.
With this architecture, the prediction of a new image (query image) will be based on the trained model and a training image (support image).
In practice, the L3IRIS team trained the HSNet~\cite{min2021hypercorrelation} from scratch with a Resnet 50 backbone
from available data to predict the content area in the input map with 512$\times$512 image input.  Because
post-processing techniques to smooth and fill edges did not improve the evaluation error, the authors kept the results
from the single trained model as the final predicted maps.

\subsection{Results and discussion}
\begin{figure}[tb]
    \centering
    \begin{subfigure}{0.39\textwidth}
        \centering\begin{tabular}{clc}
            \toprule
            Rank & Team & Hausdorff 95 (pix.) $\downarrow$\\
            \midrule
            1 & UWB & 19\\
            2 & CMM & 85\\
            3 & IRISA & 112\\
            4 & L3IRIS & 126\\
            \bottomrule
        \end{tabular}
    \end{subfigure}
    \hfill
    \begin{subfigure}{0.59\textwidth}
        \centering
        \includegraphics[width=\linewidth,trim={0 20px 5px 0},clip]{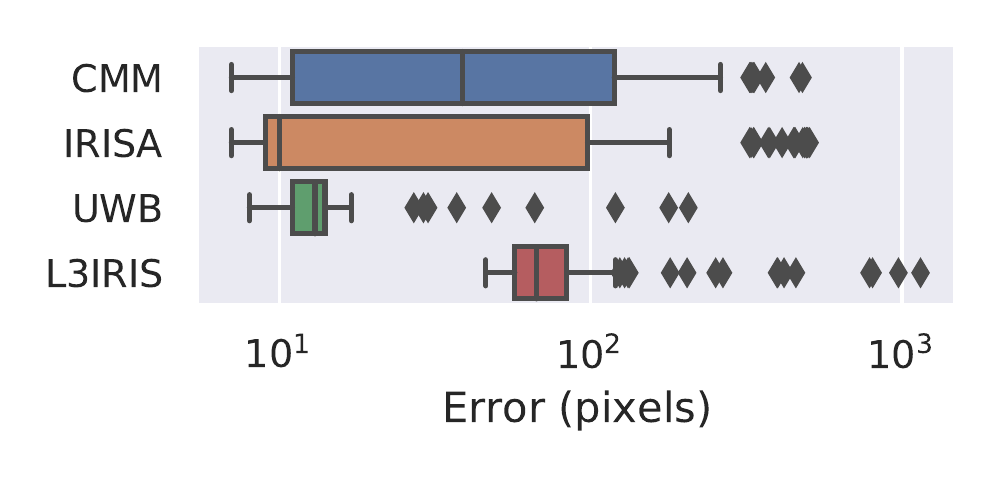}
    \end{subfigure}
    \caption{Final Hausdorff 95 errors for Task~2 (\emph{left}) --- values range from 0 pixels (best) to arbitrarily large values,
    and error distribution for all test images (\emph{right}).}
    \label{fig:task2-results}
\end{figure}
Results and ranking of each submitted method is given in~\cref{fig:task2-results}:
table on the left summarizes the Hausdorff 95 indicators we computed
and the plot on the right illustrates the error distribution (in log scale) for all test images.
The L3IRIS method, based on a single FCN leveraging recent few-shot learning technique produces overly smoothed masks.
The IRISA method, based on a robust line segment detector embedded in a rule-based system, was penalized by unseen frame configurations.
However, it produced very accurate results for know configurations.
The CMM approach, based on morphological processing, also produced very accurate frame detection and generalized better.
The remaining errors are due to missed regions considered as background close to the boundary.
The UWB winning method, finally, efficiently combines a coarse map content detection thanks to a deep network, with a re-adjustment of detected boundaries using a recursive Otsu binarization and some morphological post-processing.

\section{Task~3: Locate Graticule Lines Intersections}
\begin{figure}[tbp]
    \centering
    \includegraphics[width=0.85\textwidth]{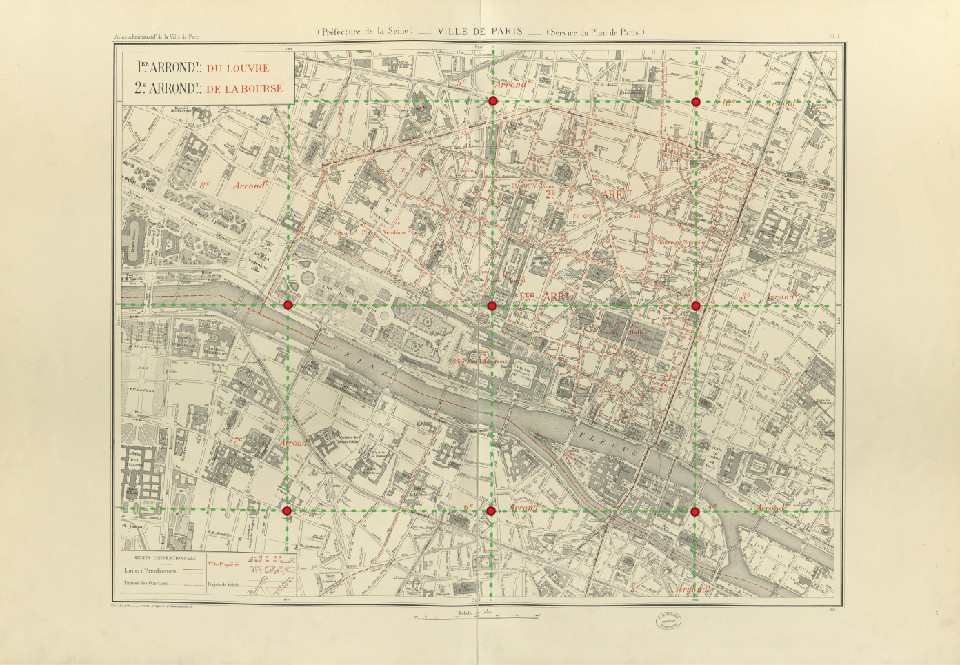}
    \caption{Illustration of expected outputs for Task~3: dashed green lines are the graticule lines and red dots are the intersections points to locate.}
    \label{fig:task3-output-large}
\end{figure}

This task consists in locating the intersection points of graticule lines.
Graticule lines are lines indicating the North/South/East/West major coordinates in the map.
They are drawn every 1000 meters in each direction and overlap with the rest of the map content.
Their intersections are very useful to geo-reference the map image, i.e. for projecting map content in a modern geographical coordinate reference system.
Given the image of a complete map sheet, participants had to locate the intersection points of such lines, as illustrated in \cref{fig:task3-output-large}.

\subsection{Dataset}
The inputs for this task are exactly the same as Task~2.
Expected output for this task is, for each input image, the list intersection coordinates.
The dataset is separated exactly as for Task~2 in terms of train, validation and test sets.

\subsection{Evaluation Protocol}
For each map sheet, we compared the predicted coordinates with the expected ones.
We used an indicator which considers detection and localization accuracy simultaneously, like what we did for Task~1.
A predicted point was considered as a correct detection if it was the closest predicted point of a ground truth (expected) point and if the distance between the expected point and the predicted one is smaller than a given threshold.
We considered all possible thresholds between 0 and 50 pixels, which roughly corresponds to 20 meters for these map images, and is an upper limit over which the registration would be seriously disrupted.
In practice, we computed for each map sheet and for each possible threshold the number of correct predictions, the number of incorrect ones and the number of expected elements.
This allowed us to plot the $F_{\beta}$ score vs threshold curve for a range of thresholds.
We set $\beta=0.5$ to weights recall lower than precision because for this task it takes several good detections to correct a wrong one in the final registration.
The area under this ``$F_{0.5}$ score vs threshold'' curve was used as performance indicator;
such indicator blends two indicators: point detection and spatial accuracy.
Finally, we computed the average of the measures for all individual map images to produce a global indicator.
The resulting measure is a value between 0 (worst) and 1 (best).

\subsection{Method descriptions}

\subsubsection{UWB --- \emph{Winning Method}}
This method is based on three main steps.
\begin{enumerate*}[label=(\roman*)]
    \item First, a binary image is generated using the recursive Otsu approach described in \cref{sec:task2-uwb}.
    This image is then masked using the map content area predicted for Task~2.
    \item Graticule lines are then detected a Hough Line Transform.
    While the accumulator bins contain a lot of noise, the following heuristics were used to filter the candidates: graticule lines are assumed to be straight, to be equally spaced and either parallel or perpendicular, and there should be at least four lines in each image.
    To enable finding, filtering, correcting, fixing and rating peak groups that represents each graticule line in Hough accumulator, each candidate contains information about its rating, angle and distance between lines.
    Rating and distance information is used to select the best configuration.
    \item Intersections are finally coarsely estimated from the intersections between Hough lines, then corrected and filtered using the predicted mask from the Task~2, using some template matching with a cross rotated by the corresponding angle.
\end{enumerate*}
The approach does not require any learning and the parameters are calibrated using both train and validation subsets.

\subsubsection{IRISA}
This method is based on two main steps.
\begin{enumerate*}[label=(\roman*)]
    \item The same line segment detector as the one used by the IRISA team for Task~2 (\cref{sec:task2-irisa}) is used to detect candidates.
    This results in a large amount of false positives with segments detected in many map objects.
    \item The DMOS rule grammar system~\cite{lemaitre08} is used to efficiently filter candidates using a dedicated set of rules.
    Such rules enable the integration of constraints like perpendicularity or regular spacing between the lines,
    and the exploration of the hypothesis space thanks to the efficient back-tracking of the underlying logical programming framework.
\end{enumerate*}

\subsubsection{CMM}
This method combines morphological processing and the Radon transform. Morphological operators can be disturbed by noise disconnecting long lines while the Radon transform can integrate the contribution of short segments that may correspond to a fine texture.
The process consists in the following six steps.
\begin{enumerate*}[label=(\roman*)]
\item First, the image is sub-sampled by a factor 10 in order to speed up the process and to make it more robust to potential line discontinuities. An erosion of size 10 is applied before sub-sampling to preserve black lines.
\item Then the frame is detected. Oriented gradients combined with morphological filters are used for this purpose, inspired by the method proposed in~\cite{hernandez2009morphological} for building fa\c{c}ade analysis.
\item Line directions are then found. Directional closings from 0 to 30 degrees with a step of 1 degree, followed by black top-hat and Otsu threshold are applied. The angle leading to the longest detected line is selected.
\item The Radon transform of the previous black top-hat at the selected direction and its orthogonal are computed. Line locations correspond to the maxima of the Radon transform.
\item Lines are equally spaced in the map. The period of this grid is obtained as the maximum of the autocorrelation of the Radon transform. Equidistant lines are added on the whole map, whether they have been detected by the Radon transform or not. Applied to both directions, this generates the graticule lines.
\item Finally a refinement is applied to localize precisely the line intersections at the full scale. 
In the neighborhood of each detected intersection, a closing is applied with a cross structuring element (with lines of length 20) at the previously detected orientations. The minimum of the resulting image provides the precise location of the intersection. If the contrast of the resulting closing is too low ($< 10$), the point is inferred from the other points that were successfully found in the refinement step on the corresponding intersecting lines.
\end{enumerate*}

\subsubsection{L3IRIS}
This method is based on a deep intersection detector followed by a filtering technique based on a Hough transform.
The process consists in the following five steps.
\begin{enumerate*}[label=(\roman*)]
    \item First, cross signs (all the points that are similar to the target points) are detected using a U-Net model~\cite{ronneberger2015u} for point segmentation, trained with the 26 training images.
    The ground truth was generated by drawing a circle at each point's location over a zero background.
    The resulting model detects many points, among which the target points.
    \item From these candidates, a Hough transform is used (with additional heuristics) to detect graticule lines.
    The parameters of the Hough transform are selected automatically based on the number of final points in the last step.
    \item A clustering algorithm is then used to select line candidates forming a regular grid (parallel and orthogonal lines).
    \item Finally, intersections between these lines are filtered using the predicted mask from Task~2 (\cref{sec:task2-l3iris}) to get the final set of points. 
\end{enumerate*}

\subsection{Results and discussion}
\begin{figure}[tb]
    \centering
    \begin{subfigure}{0.49\textwidth}
        \centering\begin{tabular}{clc}
            \toprule
            Rank & Team & Detection score (\%) $\uparrow$\\
            \midrule
            1 & UWB & 92.5\\
            2 & IRISA & 89.2\\
            3 & CMM & 86.6\\
            4 & L3IRIS & 73.6\\
            \bottomrule
        \end{tabular}
    \end{subfigure}
    \hfill
    \begin{subfigure}{0.49\textwidth}
        \centering
        \includegraphics[width=\linewidth,trim={0px 75px 15px 90px},clip]{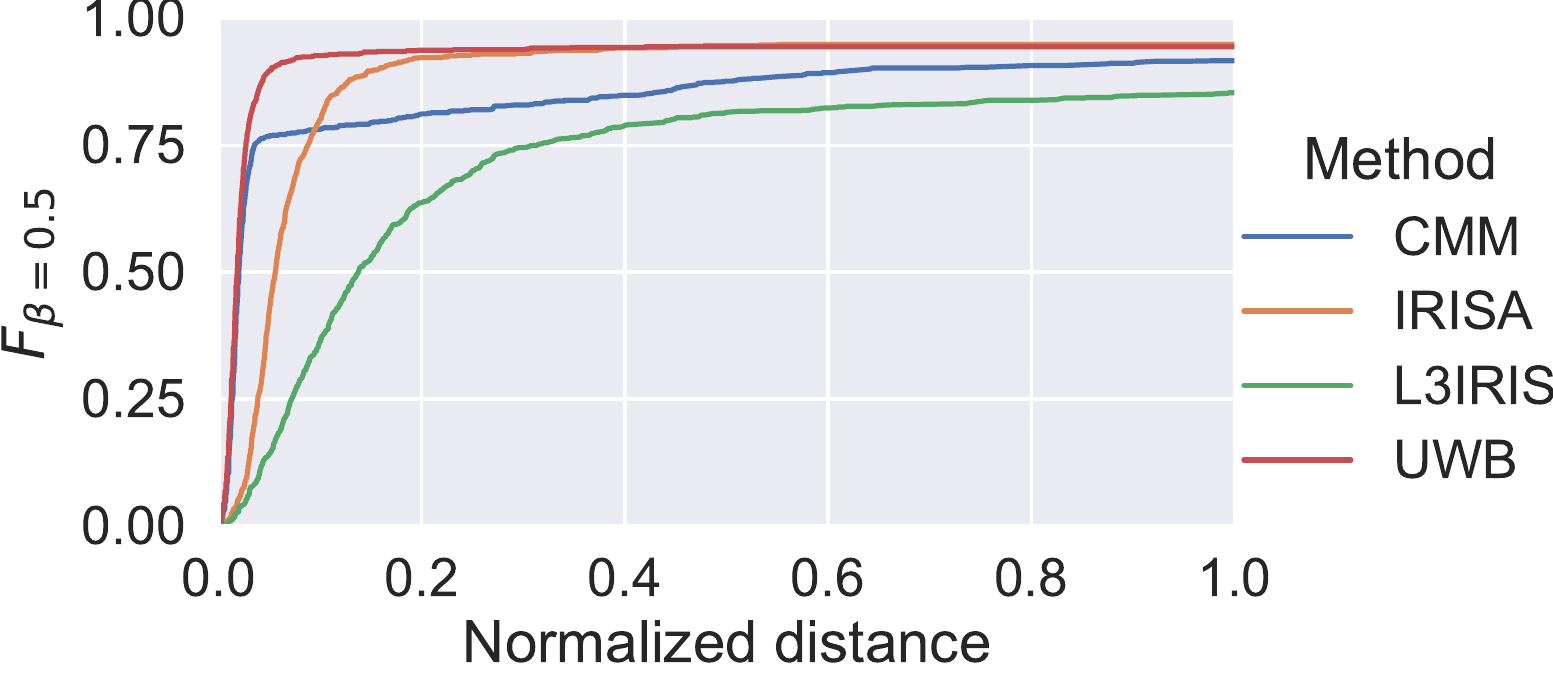}
    \end{subfigure}
    \caption{Final detection score for Task~3 (\emph{left}) --- score ranges from 0 (worst) to 100 (best),
    and plot of the $F_{\beta=0.5}$ value for each distance threshold (\emph{right}) --- normalized by the maximum distance of 50 pixels.}
    \label{fig:task3-results}
\end{figure}
Results and ranking of each submitted method is given in~\cref{fig:task3-results}:
table on the left summarizes the detection score we computed
and the plot on the right illustrates the evolution of the retrieval $F_{\beta}$ score for each possible normalized distance threshold.
The L3IRIS approach is penalized by a lack of local confirmation of intersection hypothesis, leading to false positives, misses, and sub-optimal location.
The CMM approach produced very accurate detections thanks to morphological filters, but the Radon transform seems to generate extra hypothesis or discard groups of them and therefore has some stability issues.
The IRISA approach performed globally very well thanks to is robust line segment detector and its rule-based system, despite some false positives.
The UWB winning approach, finally, efficiently combined a binary preprocessing with a coarse Hough transform and a final refinement leading to superior accuracy.

\section{Conclusion}
This competition succeeded in advancing the state of the art on historical atlas vectorization,
and \textbf{we thank all participants for their great submissions}.
Shape extraction (Task~1) still require some progress to automate the process completely.
Map content detection (Task~2) and graticule line detection (Task~3) are almost solved by proposed approaches.
Future work will need to improve on shape detection,
and start to tackle shape classification and text recognition.%

\bibliographystyle{splncs04}
\bibliography{references}

\begin{thebibliography}{10}
\providecommand{\url}[1]{\texttt{#1}}
\providecommand{\urlprefix}{URL }
\providecommand{\doi}[1]{https://doi.org/#1}

\bibitem{chronseg}
Baloun, J., Král, P., Lenc, L.: Chronseg: Novel dataset for segmentation of
  handwritten historical chronicles. In: Proc. of the 13th International
  Conference on Agents and Artificial Intelligence ({ICAART}). pp. 314--322
  (2021)

\bibitem{augmentor}
Bloice, M.D., Roth, P.M., Holzinger, A.: {Biomedical image augmentation using
  Augmentor}. Bioinformatics  \textbf{35}(21),  4522--4524 (2019)

\bibitem{blusseau18tropical}
Blusseau, S., Velasco-Forero, S., Angulo, J., Bloch, I.: Tropical and
  morphological operators for signal processing on graphs. In: Proc. of the
  25th IEEE International Conference on Image Processing (ICIP). pp. 1198--1202
  (2018)

\bibitem{chazalon21.icdar}
Chazalon, J., Carlinet, E.: Revisiting the coco panoptic metric to enable
  visual and qualitative analysis of historical map instance segmentation. In:
  16th International Conference on Document Analysis and Recognition (ICDAR)
  (2021), to appear

\bibitem{grandvalet2004semi}
Grandvalet, Y., Bengio, Y.: Semi-supervised learning by entropy minimization.
  In: Proc. of the 17th International Conference on Neural Information
  Processing Systems (NIPS). pp. 529--536 (2004)

\bibitem{hernandez2009morphological}
Hern{\'a}ndez, J., Marcotegui, B.: Morphological segmentation of building
  fa{\c{c}}ade images. In: Proc. of the 16th International Conference on Image
  Processing (ICIP). pp. 4029--4032. IEEE (2009)

\bibitem{huang2017densely}
Huang, G., Liu, Z., Van Der~Maaten, L., Weinberger, K.Q.: Densely connected
  convolutional networks. In: Proc. of the IEEE Conference on Computer Vision
  and Pattern Recognition (CVPR). pp. 4700--4708 (2017)

\bibitem{kirillov2019panoptic}
Kirillov, A., He, K., Girshick, R., Rother, C., Doll{\'a}r, P.: Panoptic
  segmentation. In: Proc. of the IEEE Conference on Computer Vision and Pattern
  Recognition (CVPR). pp. 9404--9413 (2019)

\bibitem{lemaitre08}
Lemaitre, A., Camillerapp, J., Co{\"u}asnon, B.: Multiresolution cooperation
  makes easier document structure recognition. International Journal on
  Document Analysis and Recognition (IJDAR)  \textbf{11}(2),  97--109 (2008)

\bibitem{leplumey95}
Leplumey, I., Camillerapp, J., Queguiner, C.: Kalman filter contributions
  towards document segmentation. In: International Conference on Document
  Analysis and Recognition (ICDAR). p. 765–769 (1995)

\bibitem{min2021hypercorrelation}
Min, J., Kang, D., Cho, M.: Hypercorrelation squeeze for few-shot segmentation.
  arXiv preprint arXiv:2104.01538  (2021)

\bibitem{nguyen2020learning}
Nguyen, N., Rigaud, C., Revel, A., Burie, J.: A learning approach with
  incomplete pixel-level labels for deep neural networks. Neural Networks
  \textbf{130},  111--125 (2020)

\bibitem{nina2011recursive}
Nina, O., Morse, B., Barrett, W.: A recursive otsu thresholding method for
  scanned document binarization. In: 2011 IEEE Workshop on Applications of
  Computer Vision (WACV). pp. 307--314. IEEE (2011)

\bibitem{ronneberger2015u}
Ronneberger, O., Fischer, P., Brox, T.: U-net: Convolutional networks for
  biomedical image segmentation. In: International Conference on Medical Image
  Computing and Computer-Assisted Intervention (MICCAI). pp. 234--241. Springer
  (2015)

\end{thebibliography}
\end{document}